\documentclass[10pt,twocolumn,letterpaper]{article}

\usepackage{cvpr}
\usepackage{times}
\usepackage{epsfig}
\usepackage{multirow}
\usepackage{graphicx}
\usepackage{amsmath}
\usepackage{booktabs}
\usepackage{amssymb}
\usepackage{enumerate}
\usepackage{authblk}

\usepackage[breaklinks=true,bookmarks=false]{hyperref}

\cvprfinalcopy 

\def\cvprPaperID{****} 

\begin{document}

\title{DBP: Discrimination Based Block-Level Pruning for Deep Model Acceleration}

\author[1,2]{Wenxiao Wang}
\author[1]{Shuai Zhao}
\author[1]{Minghao Chen}
\author[1]{Jinming Hu}
\author[1,2]{Deng Cai\thanks{Corresponding author}}
\author[1]{Haifeng Liu}
\affil[1]{State Key Lab of CAD\&CG, Zhejiang University, Hangzhou, China}
\affil[2]{Fabu Inc., Hangzhou, China}

\maketitle

\begin{abstract}
Neural network pruning is one of the most popular methods of accelerating the inference of deep convolutional neural networks (CNNs). The dominant pruning methods, filter-level pruning methods, evaluate their performance through the reduction ratio of computations and deem that a higher reduction ratio of computations is equivalent to a higher acceleration ratio in terms of inference time. However, we argue that they are not equivalent if parallel computing is considered. Given that filter-level pruning only prunes \textbf{filters} in layers and computations in a layer usually run in parallel, most computations reduced by filter-level pruning usually run in parallel with the un-reduced ones. Thus, the acceleration ratio of filter-level pruning is limited. To get a higher acceleration ratio, it is better to prune redundant \textbf{layers} because computations of different layers cannot run in parallel. In this paper, we propose our Discrimination based Block-level Pruning method (DBP). Specifically, DBP takes a sequence of consecutive layers (\eg, \textit{Conv-BN-ReLu}) as a block and removes redundant blocks according to the discrimination of their output features. As a result, DBP achieves a considerable acceleration ratio by reducing the depth of CNNs. Extensive experiments show that DBP has surpassed state-of-the-art filter-level pruning methods in both accuracy and acceleration ratio. Our code will be made available soon.
\end{abstract}


\section{Introduction} \label{intro}

Deep convolutional neural networks (CNNs) have achieved great success in many computer vision tasks such as image classification~\cite{DBLP:conf/cvpr/HeZRS16,DBLP:conf/cvpr/HuangLMW17}, object detection~\cite{DBLP:conf/eccv/LiuAESRFB16,DBLP:conf/cvpr/RedmonDGF16}, and semantic segmentation~\cite{DBLP:conf/iccv/HeGDG17,DBLP:journals/corr/ChenPSA17}. However, the enormous computational cost of CNNs makes it very slow to run the model on resources-constrained devices such as mobile phones. Thus, it is essential to reduce computational cost and accelerate the inference of CNNs before deployment.


\begin{figure}[t]
\begin{center}
\includegraphics[width=1.0\linewidth]{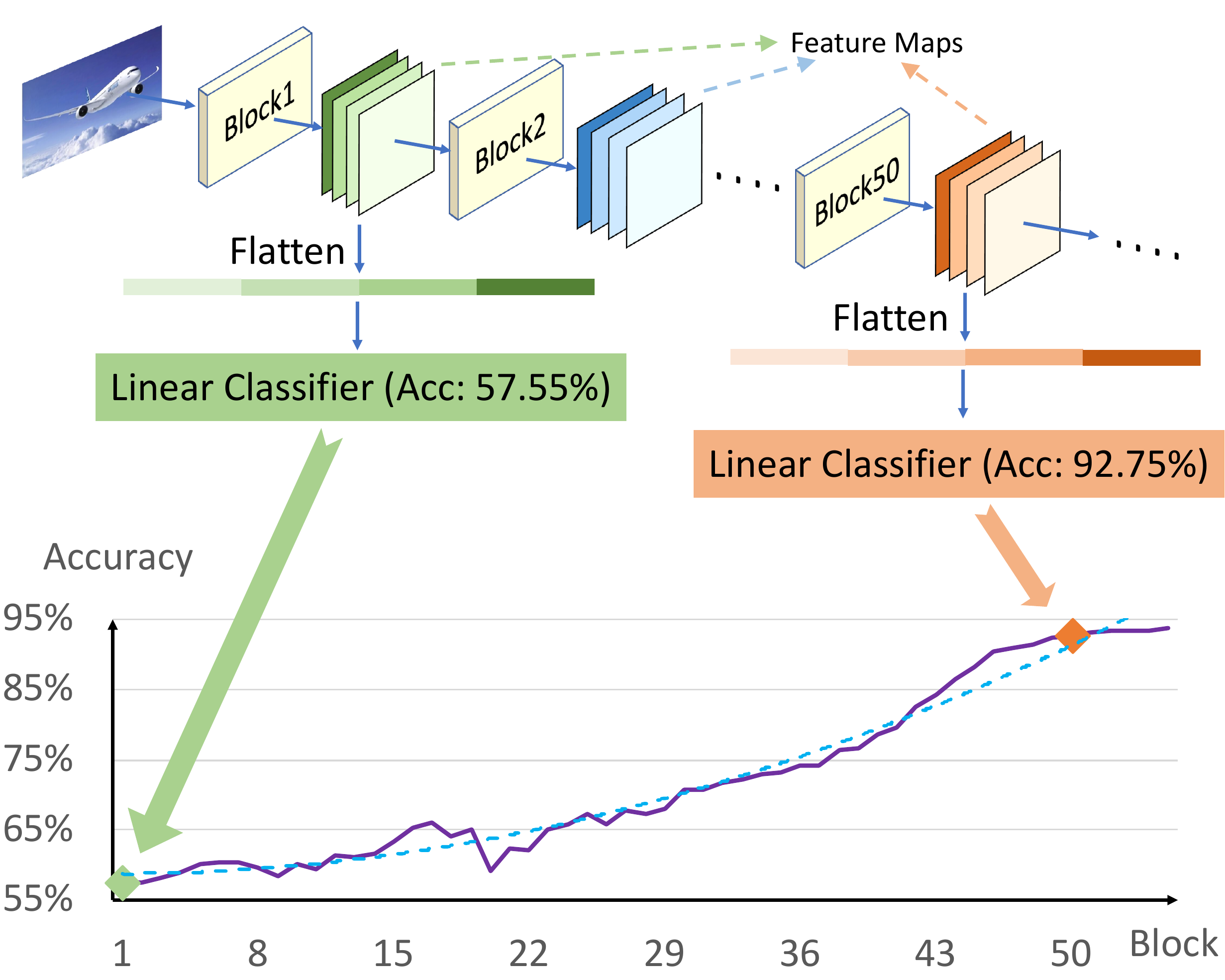}
\end{center}
\label{fig:intro}
\caption{The relations between the block depth and features' discrimination. The top part of the figure shows part of ResNet110~\cite{DBLP:conf/cvpr/HeZRS16} trained on CIFAR10~\cite{krizhevsky2009learning}. The residual blocks in ResNet110 are marked as Block1, Block2, \etc in order. We flatten the output features of each block and evaluate their discrimination through the accuracy of the linear classifiers after them. The ``Acc'' represents the accuracy on CIFAR10. We plot the accuracys of all classifiers as the solid purple line in the bottom figure and plot the trend using the blue dashed line. In general, the features' discrimination ascends as the block goes deeper, but it descends at some blocks.}
\end{figure}

Neural network pruning~\cite{han2015deep,dong2017learning,liu2017learning,he2019filter,wang2019cop} is one of the most popular model acceleration methods. It removes redundant weights or filters in CNNs to reduce computations. Most popular neural network pruning can be divided into two groups: weight-level and filter-level. Weight-level pruning~\cite{han2015deep,dong2017learning,guo2016dynamic} sets redundant weights in CNNs to zeros, making weight matrices or tensors sparse. Some previous works~\cite{liu2017learning,li2016pruning,wang2019cop} have pointed out that weight-level pruning contributes little to accelerating the inference of CNNs unless specialized libraries (such as cuSPARSE\footnote{\url{https://docs.nvidia.com/cuda/cusparse/index.html}}) are used. However, the support of these libraries on mobile devices is limited. On the other hand, filter-level pruning~\cite{liu2017learning,wang2019cop,he2019filter} removes redundant filters to reduce computations directly. Nonetheless, computations in the same layer are highly parallelized, which means most reduced computations run in parallel with the un-reduced ones. As a result, the acceleration ratio of filter-level pruning is limited.

As CNNs becomes deeper and deeper, there are many redundant layers in CNNs. Since the computations of different layers run in serial, reducing the number of layers can achieve a higher acceleration ratio than filter-level pruning methods. Thus, we propose a block-level pruning method to prune redundant layers in CNNs. Namely, we take a sequence of consecutive layers (\eg, \textit{Conv-BN-ReLu}) as a block and prune redundant blocks to reduce computations.

The key to block-level pruning is how to find redundant blocks. Motivated by~\cite{DBLP:conf/eccv/ZeilerF14,DBLP:conf/icml/BelilovskyEO19}, which proposes that the discrimination of features in CNN was enhanced block by block, we explore the discrimination of each block's output features as Figure~\ref{fig:intro}. Explicitly, we place a linear classifier after each block and test the accuracy of the classifier on a dataset. The higher the accuracy of the classifier is, the more discriminative the features are. The results show that the discrimination of features ascends as the block goes deeper, which is consistent with the conclusion of~\cite{DBLP:conf/eccv/ZeilerF14}. However, we also find that the discrimination of features ascends laxly or even descends at some blocks. Based on previous works and our observations, we assume that these blocks are redundant and can be pruned with acceptable loss.

Some algorithms also support block-level pruning.~\cite{DBLP:conf/eccv/HuangW18,DBLP:conf/nips/WenWWCL16} use a norm-based importance evaluation, which has been proven to be inappropriate in~\cite{he2019filter}.~\cite{DBLP:conf/cvpr/LinJYZCYHD19} utilizes generative adversarial learning. However, its performance is still limited, and the reason might be that generative adversarial networks are difficult to converge~\cite{2015_ALec}.

Extensive experiments show that DBP achieves a higher acceleration ratio as well as higher accuracy than state-of-the-art filter-level pruning and block-level pruning. Additionally, we also compare DBP with knowledge distillation because it can also yield shallow models that achieve high accuracy. Experiments show that DBP has surpassed the state-of-the-art knowledge distillation based model acceleration methods.

It is worthwhile to highlight our contributions:
\begin{itemize}
\item We analyze the reason for the limited acceleration ratio of filter-level pruning and propose that block-level pruning avoids the problem well.

\item We propose a discrimination based criterion to observe the redundant blocks in CNNs.

\item Extensive experiments show that DBP has surpassed state-of-the-art block-level pruning, filter-level pruning, and knowledge distillation methods in both accuracy and acceleration ratio.


\end{itemize}

\section{Related works}

Neural network pruning\cite{han2015deep,li2016pruning,liu2017learning,wang2019cop,he2019filter} and knowledge distillation~\cite{hinton2015distilling,DBLP:conf/aaai/ZhouFCBZG18,DBLP:conf/aaai/HeoLY019,DBLP:journals/corr/abs-1907-09682,DBLP:conf/eccv/BelagiannisFG18} have shown their effectiveness in accelerating the inference of deep CNNs.

\paragraph{Neural network pruning} Neural network pruning devotes to removing redundant weights in networks to accelerate the inference, and it can be divided into three groups: weight-level, filter-level, and block-level: \textbf{Weight-level} pruning observes redundant weights in filters and set them to zeros. For example,~\cite{han2015deep} proposes weights with small absolute value can be set to zeros without loss. Weight-level pruning makes weights tensors sparse, and it can accelerate the inference of CNNs with the help of specialized libraries (\eg, cuSPARSE). Unfortunately, the support of these specialized libraries on mobile devices is limited. \textbf{Filter-level} pruning solves the problem by pruning unimportant filters in CNNs and reduce computations directly. The key to filter-level pruning is how to define the importance of filters. For example,~\cite{li2016pruning,liu2017learning} take advantage of the idea of~\cite{han2015deep} and evaluate the importance of filters according to their amplitude.~\cite{wang2019cop,he2019filter} considers the interrelations among filters when evaluating importance. Generally speaking, filter-level pruning has achieved great success, but, as we have discussed above, its acceleration ratio is still limited because of parallel computing. \textbf{Block-level} pruning takes a sequence of consecutive layers as a block and remove redundant blocks to reduce the computations in CNNs. Considering that computations in different layers run in serial, pruning more blocks must mean a higher acceleration ratio.~\cite{DBLP:conf/nips/WenWWCL16,DBLP:conf/eccv/HuangW18,DBLP:conf/cvpr/LinJYZCYHD19} all support block-level pruning. Specifically,~\cite{DBLP:conf/nips/WenWWCL16,DBLP:conf/eccv/HuangW18} observe redundant blocks and filters through norm-based criterion, and GAL~\cite{DBLP:conf/cvpr/LinJYZCYHD19} utilizes generative adversarial networks (GAN) to prune blocks and filters simultaneously. However, the norm-based criterion has been shown to be inappropriate in~\cite{he2019filter}, and GAN is hard to converge.

\paragraph{Knowledge distillation} The main idea of knowledge distillation (KD) is to transfer knowledge from a trained teacher model to an un-trained student model, so KD helps on many tasks such as transfer learning, few-shot learning, and model acceleration. When working with model acceleration, KD transfers knowledge from a large teacher model to a small student model, and the key to KD is what knowledge should be transferred to the student.~\cite{hinton2015distilling,DBLP:conf/nips/BaC14} take the output of the last layer as the knowledge and induce soft targets\cite{hinton2015distilling} and mimic loss~\cite{DBLP:conf/nips/BaC14} to transfer the knowledge. Other algorithms~\cite{DBLP:conf/aaai/ZhouFCBZG18,DBLP:conf/aaai/HeoLY019,DBLP:journals/corr/abs-1907-09682} explore the knowledge from intermediate layers and combine them with soft targets and mimic loss to achieve higher accuracy. DBP also takes advantage of mimic loss when fine-tuning the model to advance its performance.

There are still many other methods to accelerate the inference of deep CNNs, such as weight quantization\cite{DBLP:conf/nips/CourbariauxBD15,DBLP:journals/corr/LiL16} and compact network design\cite{DBLP:conf/cvpr/ZhangZLS18,DBLP:conf/cvpr/SandlerHZZC18}. Most of them can be combined with pruning and knowledge distillation for further acceleration.

\section{Algorithm}

\subsection{Overview}

Deep CNNs consist of a number of repeatable blocks, and most computations of CNNs come from these blocks. Thus, We aim to accelerate the inference of CNNs through pruning redundant blocks while keeping the accuracy of CNNs.

We first introduce how we observe and prune redundant blocks, namely our discrimination based criterion, in Section~\ref{alg:sec3}. Then the strategies of recovering the performance of the model after pruning are introduced in Section~\ref{alg:sec4}. In Section~\ref{alg:sec5}, we do some customization for models with special structures. For simplicity, we use $B_i$ to represent the $i_{th}$ block in a CNN.

\begin{figure}[t]
\begin{center}
\includegraphics[width=1.0\linewidth]{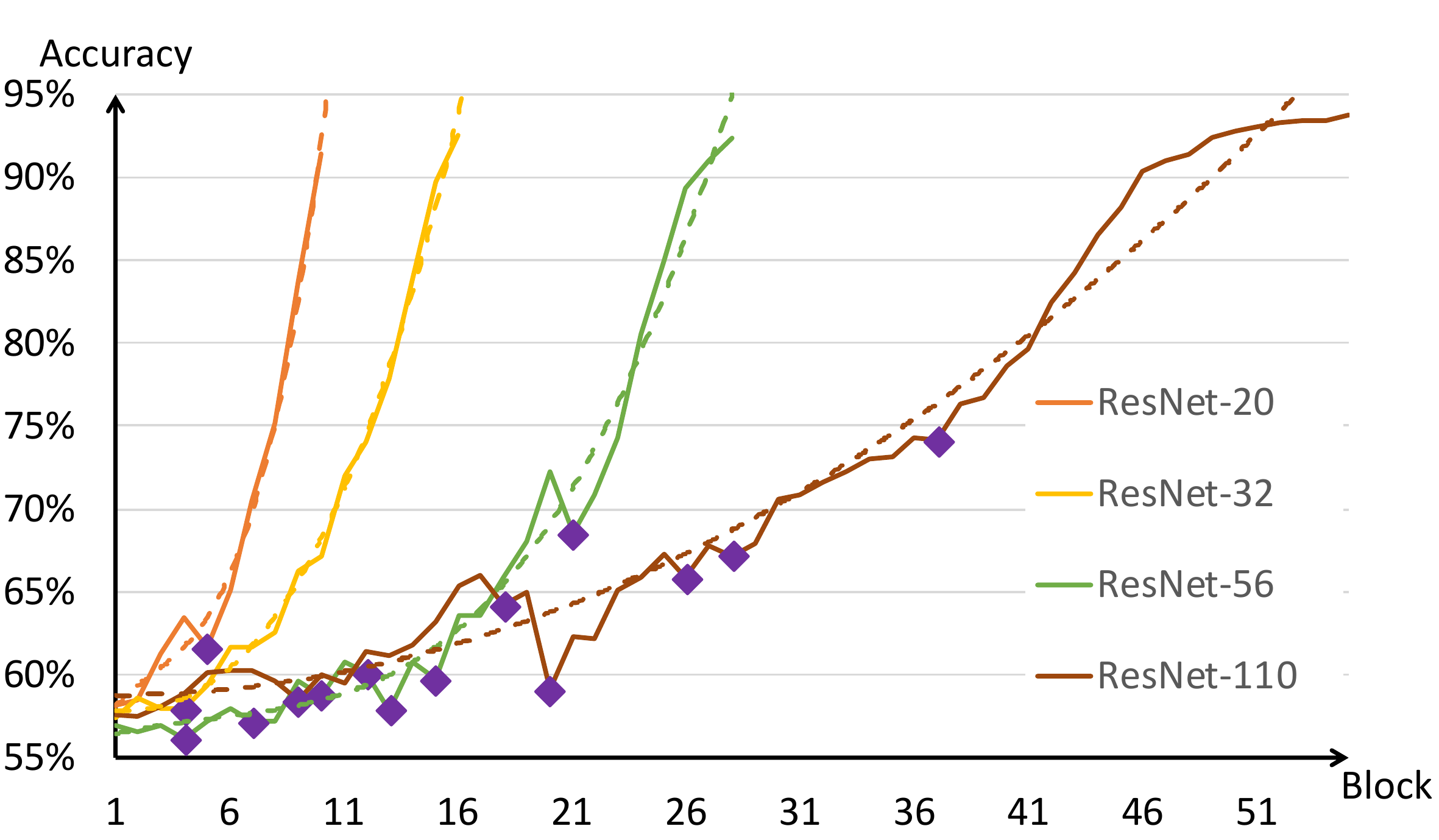}
\end{center}
\caption{The discrimination of each block's output features. The solid lines represent the accuracy of linear classifiers after each block, and dash lines are their polynomial trendlines. Generally speaking, the accuracy ascends as the block goes deeper. However, we also find: 1) The accuracy descends (at the purple markers) at some blocks no matter how deep the model is. 2) The discrimination of features in a shallow model always ascends faster than that in a deep one.}
\label{fig:trend}
\end{figure}

\subsection{Discrimination based criterion} \label{alg:sec3}

\cite{DBLP:conf/eccv/ZeilerF14} has explored the discrimination of output features of each block in CNNs by placing a linear classifier after each block and find that the discrimination of features ascends as the block goes deeper.~\cite{DBLP:conf/icml/BelilovskyEO19} shows that training a deep CNN on ImageNet block by block generates a model whose accuracy is as high as an end-to-end-trained model. Though there has not been a reasonable explanation to these phenomena, these experiments have shown that the performance of deep CNNs is gradually enhanced block by block.  

Motivated by~\cite{DBLP:conf/eccv/ZeilerF14,DBLP:conf/icml/BelilovskyEO19}, we further explore the discrimination of each block's output features in CNNs. We use a fully-connected layer as our linear classifier and place it after each block of a trained CNN. We fix the weights of the CNN and train these classifiers. Consequently, it is natural to take the accuracy of these classifiers as the discrimination of blocks' output features, and the higher the accuracy is, the more discriminative the features are. We experiment on different depths of CNNs, and the results are in Figure~\ref{fig:trend}. The accuracy ascends as the block goes deeper, which is consistent with the conclusion in~\cite{DBLP:conf/eccv/ZeilerF14}. Moreover, we also find another two interesting phenomena and propose our assumptions:

\begin{table}[t]
\begin{center}
\begin{tabular}{c|c}
    \toprule
    Removed & AC(\%) \\
    \midrule
    Degraded Block & -0.45$\pm$0.24  \\ 
    \midrule
    Upgraded Block & -0.97$\pm$0.34\\ 
    \bottomrule
\end{tabular}
\end{center}
\caption{The influence of removing a degraded or upgraded block on the accuracy changes (AC) of ResNet110. We explore the contribution of a block by removing it and checking the accuracy changes of the whole model on CIFAR10. As we see, removing a degraded block induces less accuracy loss than removing an upgraded block.}
\label{tab:alg}
\end{table}

\paragraph{Degradation} The accuracy of the classifier descends at some blocks, which indicates that their extracted features are confusing and degrade the discrimination of output features. We call them \textit{degraded blocks} and call other blocks \textit{upgraded blocks}. We explore each block's contribution to the model by removing it and check the changes of the model's accuracy. As Table~\ref{tab:alg} shows, compared with an upgraded block, removing a degraded block has limited influence on the performance of the model, which means the weights of degraded blocks have less contribution to enhancing the performance of the whole model. As a result, degraded blocks can be pruned with acceptable loss.

\paragraph{Slowness} The trend line of a shallow model is always steeper than that of a deeper one. In other words, the deeper the model is, the slower the discrimination of its features ascends. We assume that it is because many blocks in deep models contribute little to ascending the discrimination of the features, and these blocks can also be pruned with acceptable loss.

With these two assumptions, we observe and prune redundant blocks with the following steps:

\begin{enumerate}

\item Place a fully-connected layer as a linear classifier after each block of a CNN and train these classifiers.

\item Compute the contributions of all blocks. Without loss of generality, we define the contribution of block $B_i$ as $ACC_{B_i} - ACC_{B_{i-1}}$, in which $ACC_{B_i}$ means the accuracy of the linear classifier after block $B_i$.

\item Find the part of blocks which have the least contributions and prune them.

\end{enumerate}

\subsection{Performance recovery}  \label{alg:sec4}
\label{method:fine-tune}

Pruning blocks will hurt the performance of the model, so we fine-tune the model after pruning to recover its performance. We take advantage of some techniques from knowledge distillation and filter-level pruning during the process of fine-tuning: 

\paragraph{Mimic loss} It is a common technique in knowledge distillation that using the mimic loss~\cite{DBLP:conf/nips/BaC14,DBLP:conf/aaai/ZhouFCBZG18,DBLP:conf/cvpr/LinJYZCYHD19} to advance the accuracy of the model. Specifically,~\cite{DBLP:conf/nips/BaC14} finds that forcing the small (student) model to mimic the last layer's output of the huge (teacher) model helps to improve the performance of the small model and proposes its $\ell2$-based mimic loss. We follow the manner of~\cite{DBLP:conf/nips/BaC14} and re-define our loss function during the fine-tuning process as Equation~\ref{equ:method0}, in which $\alpha$ is a hyper-parameter, $l_T$ and $l_S$ are the logits (the output before activations) of the unpruned model and pruned model respectively, $p_i$ is the one-hot label of ground-truth and $q = softmax(l_S)$. The first term in the equation is the mimic loss, and the second term is commonly used~\cite{DBLP:conf/cvpr/HeZRS16,DBLP:conf/cvpr/HuangLMW17} cross-entropy loss in classification.

\begin{equation}
\label{equ:method0}
\begin{aligned}
loss = \alpha * ||l_{T} - l_S||_2^2 + \sum_i -p_i \, \log \, q_i
\end{aligned}
\end{equation}

\paragraph{Iterative pruning} Iterative pruning has shown to help improve the performance of pruning in many papers~\cite{DBLP:conf/iccv/LuoWL17,DBLP:conf/iclr/0022KDSG17,DBLP:conf/iclr/MolchanovTKAK17}. Explicitly, they prune redundant filters layer by layer and fine-tune the model after pruning each layer because pruning all layers once may lead to invocatable accuracy loss. Besides, research on knowledge distillation~\cite{DBLP:journals/corr/abs-1902-03393} also proposes that it will hurt the performance of the student model if the teacher model is much more complicated than the student. Motivated by them, we also prune redundant blocks in an iterative manner. Specifically, we use a hyper-parameter $\beta$ to control the ratio of blocks to prune each time and repeat pruning and fine-tuning steps until we get a model as small as we need. Considering that there is a tradeoff between accuracy and acceleration ratio, iterative pruning has another advantage that it would generate some intermediate models, and users can stop pruning once they meet a satisfying one.

\begin{figure}[t]
\begin{center}
\includegraphics[width=1.0\linewidth]{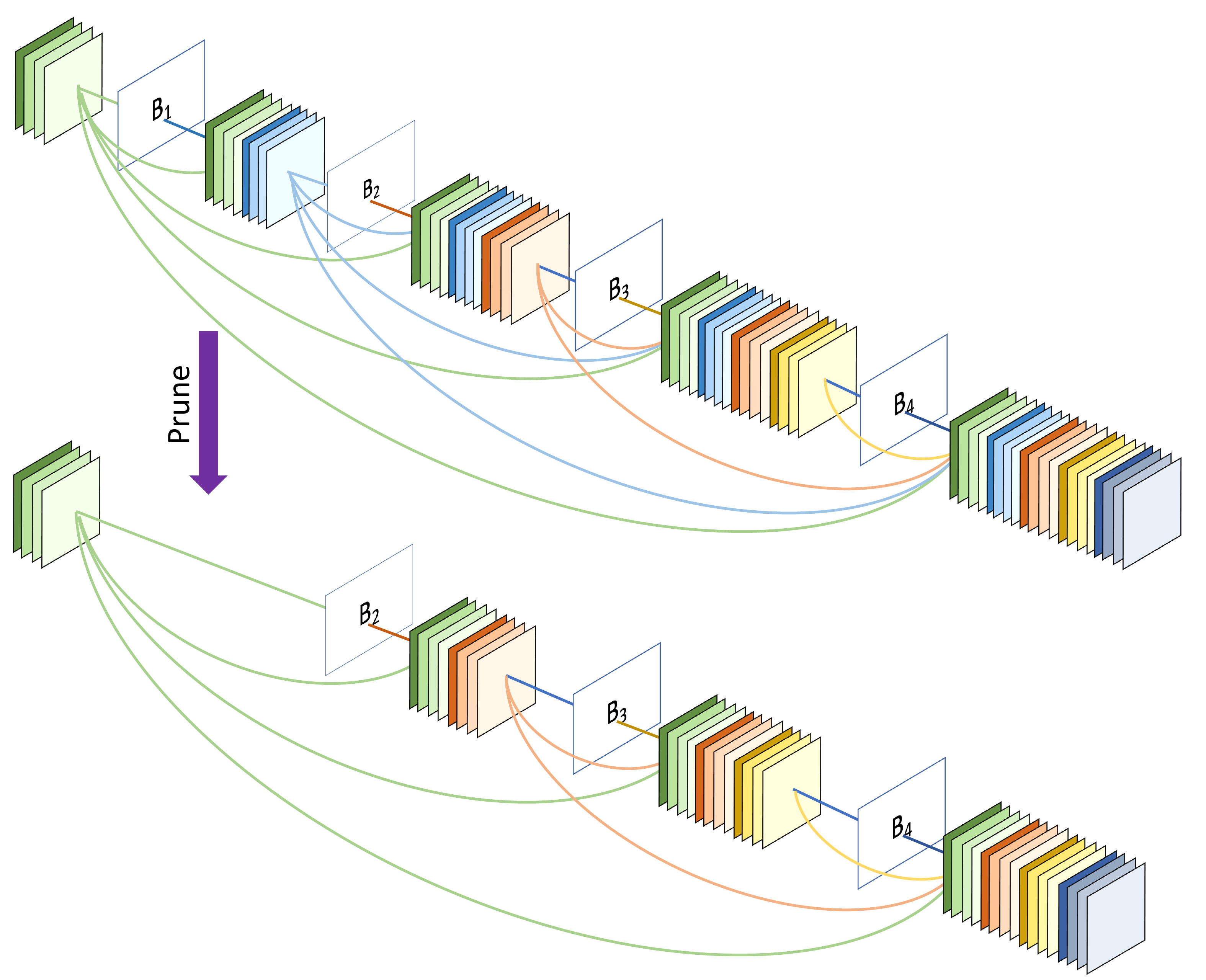}
\end{center}
\caption{An example of pruning DenseNet. The colored stacked squares are feature maps. As we see, the feature maps in DenseNet are densely connected~\cite{DBLP:conf/cvpr/HuangLMW17}, and if $B_1$ is pruned, the number of input channels of blocks after $B_1$ will be reduced.}
\label{fig:densenet}
\end{figure}

\subsection{Customization} \label{alg:sec5}

\paragraph{For ResNet} \textit{Conv-BN-ReLu} is the most common block in CNNs. However, it is not appropriate to take it as the minimum pruning unit for ResNet owing to the special structure of ResNet. Specifically, pruning $B_i$ means we will take the output of $B_{i-1}$ as the input of $B_{i+1}$ directly. This requires the number of output channels of $B_{i-1}$ to be the same as the number of input channels of $B_{i+1}$. However, the bottleneck design in ResNet makes most $i$ do not satisfy the condition. As a result, we take a residual block in ResNet as the minimum pruning unit, and then most blocks can be pruned safely. Moreover,~\cite{DBLP:conf/eccv/HuangW18} also proposes that removing a residual block in ResNet does not cut off the information flow in ResNet, which is friendly to block-level pruning.

\paragraph{For DenseNet} According to the definition in DenseNet~\cite{DBLP:conf/cvpr/HuangLMW17}, each dense block contains too many layers, and it is difficult to keep the accuracy of the model once we prune one block. So, we have to re-define the block in DenseNet as a sequence of consecutive \textit{BN-ReLu-Conv}. Note that we only give ``block'' a new meaning without changing the structure of DenseNet. Consequently, like the example shown in Figure~\ref{fig:densenet}, the output of a certain block will be the input of many other blocks, and pruning a block $B_i$ means that the number of input channels of all blocks after $B_i$ will be reduced.





\section{Experiments}

\subsection{Datasets and architectures}

We experiment on two well-known classification datasets: CIFAR~\cite{krizhevsky2009learning} and ImageNet~\cite{DBLP:journals/ijcv/RussakovskyDSKS15}. CIFAR contains 50,000 images for training and 10,000 images for evaluation, and the size of each image is $32 \times 32$. CIFAR10 and CIFAR100 are two splits in CIFAR, which contain 10 and 100 classes, respectively, and we use both of them when performing experiments. ImageNet is a dataset containing 1.28 million training images and 50,000 testing images for 1000 categories of objects. The size of each image in ImageNet is $224 \times 224$.

We use two famous deep CNNs to take experiments: ResNet\cite{DBLP:conf/cvpr/HeZRS16} and DenseNet\cite{DBLP:conf/cvpr/HuangLMW17}. Both of them have very deep versions of over a hundred layers and achieve competitive results. We try these two CNNs with several different depths, and our DBP achieves great success on all of them.

\begin{table}
\begin{center}
\scalebox{0.96}{
\begin{tabular}{c|lrrrr}
    \toprule
    Dataset & Methods & Acc(\%) & Frr(\%) & Time(ms)/AR \\
    \midrule
    \multirow{10}{*}{C10}
    & R56 & \textbf{93.72} & 0.00 & 63.79/1.00 \\
    & GAL\cite{DBLP:conf/cvpr/LinJYZCYHD19} & 93.38 & 37.68 & 47.67/1.34 \\
    & GAL\cite{DBLP:conf/cvpr/LinJYZCYHD19} & 91.58 & 60.20 & 30.51/2.09 \\
    & DBP-0.5 & \textbf{93.39} & 53.41 & 29.33/2.17 \\
    & DBP-0.7 & 92.32 & \textbf{68.69} & \textbf{21.39}/\textbf{2.98} \\
    \cline{2-5}
    & R110 & \textbf{93.97} & 0.00 & 103.43/1.00 \\
    & GAL\cite{DBLP:conf/cvpr/LinJYZCYHD19} & 93.59 & 18.70 & 89.11/1.16 \\
    & GAL\cite{DBLP:conf/cvpr/LinJYZCYHD19} & 92.74 & 48.50 & 63.65/1.62 \\
    & DBP-0.4 & \textbf{93.61} & 43.92 & 69.36/1.49 \\
    & DBP-0.5 & 93.25 & \textbf{56.98} & \textbf{54.59}/\textbf{1.89} \\
    \midrule
    \multirow{6}{*}{I1000}
    & R50 & \textbf{76.15} & 0.00 & 126.97/1.00 \\
    & SSS\cite{DBLP:conf/eccv/HuangW18} & 74.18 & 31.05 & 102.11/1.24 \\
    & SSS\cite{DBLP:conf/eccv/HuangW18} & 71.82 & 43.04 & 92.31/1.38 \\
    & GAL\cite{DBLP:conf/cvpr/LinJYZCYHD19} & 71.95 & 43.04 & 90.82/1.40 \\
    & DBP-0.4 & \textbf{74.74} & 37.47 & 91.60/1.39 \\
    & DBP-0.5 & 72.44 & \textbf{49.97} & \textbf{79.35}/\textbf{1.60} \\
    \bottomrule
\end{tabular}}
\end{center}
\caption{The results compared with block-level pruning on two datasets: CIFAR10 (C10) and ImageNet (I1000). Acc and Frr mean accuracy and FLOPs-reduction-ratio, respectively. ``Time'' is the inference time per image, and ``AR'' is the corresponding acceleration ratio of the pruned model. DBP-$x$ means the pruned ratio of blocks is $x$.}
\label{tab:BP1}
\end{table}

\subsection{Compared algorithms and evaluation protocol}

We compare our DBP with three kinds of model acceleration methods: block-level pruning, knowledge distillation, and filter-level pruning. Both block-level pruning and knowledge distillation based acceleration methods reduce the number of blocks in CNNs, so it is necessary to compare with them. Besides, we also compare with filter-level pruning to highlight our considerable acceleration ratio over them, though filter-level pruning based methods are not our opponents since DBP can be combined with them for further acceleration.

The target of model acceleration is accelerating the inference as well as keeping the accuracy of models. Other papers on pruning~\cite{li2016pruning,liu2017learning,wang2019cop,he2019filter} use the accuracy of the pruned model and the reduction ratio of floating-point-operations (FLOPs) to evaluate their algorithms. We add a third evaluation protocol, namely, the acceleration ratio. It is worth recalling the definition of acceleration ratio (AR) as Equation~\ref{equ:exp0}, in which $Time_o$ and $Time_p$ represent the inference time of the original huge model and the pruned model, respectively. Acceleration ratio is the most direct metric of acceleration effect.

\begin{equation}
\label{equ:exp0}
\begin{aligned}
AR = \frac{Time_{o}}{Time_{p}}
\end{aligned}
\end{equation}

\subsection{Implementation details}

We train baseline models on CIFAR with the same settings as original papers~\cite{DBLP:conf/cvpr/HeZRS16,DBLP:conf/cvpr/HuangLMW17}. The baseline models on ImageNet are downloaded from the official website of PyTorch\footnote{\url{https://pytorch.org/}}.

The linear classifiers after each block are trained on the same dataset as training the baseline. Considering that there are only few parameters in a linear classifier, the classifiers are trained only three epochs with learning rate [0.1, 0.01, 0.001] for each epoch when experimenting on CIFAR. When experimenting on ImageNet, we train only 3,000 iterations with learning rate [0.1, 0.01, 0.001] for each 1,000 iterations to save time.

 When fine-tuning the model, we always set $\alpha$ in Equation~\ref{equ:method0} to 1.0 for all experiments and use the same data augmentation and batch size as training the baseline. With a global pruning ratio ($G$), all models are pruned three times iteratively, so the pruned ratio $\beta$ for each time is computed as $\beta = 1 - (1-G)^{\frac{1}{3}}$. Then we fine-tune the pruned model for $\frac{1}{5}$ as many epochs as training the baseline for each time of pruning. Thus, the number of fine-tuning epochs is $\frac{3}{5}$ as big as that of training the baseline.

\begin{table}[t]
\begin{center}
\scalebox{0.92}{
\begin{tabular}{c|cclr}
    \toprule
    Dataset & T-S & T-Acc(\%) & Methods & S-Acc(\%) \\
    \midrule
    
    \multirow{16}{*}{C10}
    & \multirow{6}{*}{R164-R20}
    & \multirow{6}{*}{94.16}
    & NONE & 91.74 \\
    & & & ANC$^\dagger$\cite{DBLP:conf/eccv/BelagiannisFG18} & 91.92 \\
    & & & AT\cite{DBLP:conf/iclr/ZagoruykoK17} & 91.13 \\
    & & & BSS\cite{DBLP:conf/aaai/HeoLY019} & 91.43 \\
    & & & RoLa\cite{DBLP:conf/aaai/ZhouFCBZG18} & 92.65 \\
    & & & DBP & \textbf{92.91}  \\
    \cline{2-5}
    & \multirow{6}{*}{R26-R8}
    & \multirow{6}{*}{92.55}
    & NONE & 88.80 \\
    & & & GIFT$^\dagger$\cite{DBLP:conf/cvpr/YimJBK17} & 88.70 \\
    & & & AT\cite{DBLP:conf/iclr/ZagoruykoK17} & 89.15 \\
    & & & BSS\cite{DBLP:conf/aaai/HeoLY019} & 88.42 \\
    & & & RoLa\cite{DBLP:conf/aaai/ZhouFCBZG18} & 89.53 \\
    & & & DBP & \textbf{89.69} \\
    \cline{2-5}
    & \multirow{4}{*}{D100-D40}
    & \multirow{4}{*}{94.88}
    & NONE & 92.28 \\
    & & & AT\cite{DBLP:conf/iclr/ZagoruykoK17} & 93.07 \\
    & & & BSS\cite{DBLP:conf/aaai/HeoLY019} & 92.99 \\
    & & & DBP & \textbf{93.84} \\
    \midrule

    \multirow{15}{*}{C100}
    & \multirow{4}{*}{R164-R20}
    & \multirow{4}{*}{72.32}
    & NONE & 67.69 \\
    & & & AT\cite{DBLP:conf/iclr/ZagoruykoK17} & 67.72 \\
    & & & BSS\cite{DBLP:conf/aaai/HeoLY019} & 68.66 \\
    & & & RoLa\cite{DBLP:conf/aaai/ZhouFCBZG18} & 69.45 \\
    & & & DBP & \textbf{69.67}  \\
    \cline{2-5}
    & \multirow{6}{*}{R32-R14}
    & \multirow{6}{*}{69.36}
    & NONE & 65.72 \\
    & & & GIFT$^\dagger$\cite{DBLP:conf/cvpr/YimJBK17} & 66.33 \\
    & & & AT\cite{DBLP:conf/iclr/ZagoruykoK17} & 66.69 \\
    & & & BSS\cite{DBLP:conf/aaai/HeoLY019} & 66.72 \\
    & & & RoLa\cite{DBLP:conf/aaai/ZhouFCBZG18} & 67.11 \\
    & & & DBP & \textbf{67.34} \\
    \cline{2-5}
    & \multirow{4}{*}{D100-D40}
    & \multirow{4}{*}{76.52}
    & NONE & 70.17 \\
    & & & AT\cite{DBLP:conf/iclr/ZagoruykoK17} & 70.98 \\
    & & & BSS\cite{DBLP:conf/aaai/HeoLY019} & 70.39 \\
    & & & DBP & \textbf{72.65} \\
    \midrule

    \multirow{7}{*}{I1000}
    & \multirow{4}{*}{R101-R50}
    & \multirow{4}{*}{77.37}
    & AT\cite{DBLP:conf/iclr/ZagoruykoK17} & 74.65 \\
    & & & BSS\cite{DBLP:conf/aaai/HeoLY019} & 73.66 \\
    & & & RoLa\cite{DBLP:conf/aaai/ZhouFCBZG18} & 75.02 \\
    & & & DBP & \textbf{76.97} \\
    \cline{2-5}
    & \multirow{3}{*}{D121-D65}
    & \multirow{3}{*}{74.42}
    & AT\cite{DBLP:conf/iclr/ZagoruykoK17} & 70.97 \\
    & & & BSS\cite{DBLP:conf/aaai/HeoLY019} & 68.76 \\
    & & & DBP & \textbf{71.37} \\
    \bottomrule

\end{tabular}}
\end{center}
\caption{The results compared with knowledge distillation. C10, C100, and I1000 mean CIFAR10, CIFAR100, and ImageNet datasets, respectively. T-S lists models and the depths of teachers and students, and R$x$-R$y$ (or D$x$-D$y$) means we prune ResNet (or DenseNet) from $x$ layers to $y$ layers. T-Acc and S-Acc are the accuracies of teacher and student, respectively. ``NONE'' means we train the student model from scratch without knowledge distillation. Algorithms with $\dagger$ use their manually designed student models in their papers. Others use the same student models as our DBP pruned models.}
\label{tab:KD1}
\end{table}

\subsection{Results and analysis}

\subsubsection{Results compared with block-level pruning}
\label{exp:bp}

To the best of our knowledge, there has not been any algorithm only for block-level pruning, but some algorithms such as GAL\cite{DBLP:conf/cvpr/LinJYZCYHD19} and SSS\cite{DBLP:conf/eccv/HuangW18} perform block-level and filter-level pruning simultaneously. When comparing with them, DBP just prunes redundant blocks while GAL and SSS prune both redundant blocks and filters. Notably, we use two 2.2GHz CPU to run the model with $batch\_size=1$, and we run 100 examples and take their average inference time as the result.

As shown in Table~\ref{tab:BP1}, DBP performs far better than them in both accuracy and acceleration ratio, even though we only prune redundant blocks. We ascribe our advantages over them to two aspects: 1) Some methods like SSS add an extra special designed regularizer to the loss function and train the huge model as well as generate the pruned model in one pass. Considering that the training of all blocks is constrained by the regularizer, the performance of the model will be influenced by the regularizer. In contrast, DBP never uses extra regularizer besides common $l2$ regularizer when training the huge model, so the huge model is easier to do its best. It is obvious that the good performance of huge model benefits. 2) Some methods, like GAL, utilizes generative adversarial learning to prune redundant blocks. However, the generative adversarial networks might be hard to train. On the other hand, we prune blocks through discrimination based criterion, and models are easy to converge during the fine-tuning.

\subsubsection{Results compared with knowledge distillation}

Following the manners of most knowledge distillation~\cite{DBLP:conf/eccv/BelagiannisFG18,DBLP:conf/cvpr/YimJBK17,DBLP:conf/iclr/ZagoruykoK17,DBLP:conf/aaai/HeoLY019,DBLP:conf/aaai/ZhouFCBZG18} methods, we use the same teacher and student models when compared with most algorithms. As a consequence, the acceleration ratio of them is the same as ours, and we only need to compare the accuracy of their student models. For the reason that DBP decides the architecture of the pruned model automatically while knowledge distillation uses manually designed ones, we take our pruned model as the student model of knowledge distillation. ``NONE'' trains the small model from scratch without knowledge distillation and is the benchmark of all algorithms. Note that some algorithms do not publish their code, so we only compare DBP with their results in the papers which share the same architecture, depth, and FLOPs with ours. Considering that models with the same architecture, depth, and FLOPs also share similar acceleration ratios, the comparison also makes sense.

The results are shown in Table~\ref{tab:KD1}. We experiment on different depths of ResNet and DenseNet with CIFAR and ImageNet dataset. The results show that all algorithms perform better than ``NONE'', while our DBP achieves the best performance in all experiments. It is worth highlighting that the accuracy of DBP is at least 1.5\% higher than that of other algorithms on ImageNet. We do not compare with RoLa on DenseNet because RoLa cannot be used on DenseNet directly. We ascribe our advantage over these knowledge distillation methods to two reasons: 1) Compared with those using the same small (student) models as DBP, we fine-tune the model while knowledge distillation train it from scratch, and fine-tuning a trained model has been shown to be better than training from scratch in many papers~\cite{li2016pruning,han2015deep,DBLP:conf/iclr/MolchanovTKAK17}. 2) Besides the influence of fine-tuning, DBP also observes a better student architecture than them.


\begin{table}[t]
\begin{center}
\scalebox{0.91}{
\begin{tabular}{c|lrrr}
    \toprule
    Dataset & Methods & Acc(\%) & Frr(\%) & Time(ms)/AR \\
    \midrule
    \multirow{8}{*}{C10}
    & R164 & \textbf{94.17} & 0.00 & 183.13/1.00 \\
    & Slim\cite{liu2017learning} & 94.00 & 53.69 & 127.12/1.44 \\
    & FPGM\cite{he2019filter} & 93.47 & 76.68 & 107.00/1.71 \\
    & COP\cite{wang2019cop} & 92.97 & 78.47 & 101.74/1.80 \\
    & DBP & 93.73 & 74.45 & 47.47/3.86 \\
    & DBP+Slim & 93.33 & 81.60 & 34.27/5.34 \\
    & DBP+FPGM & 93.22 & 82.98 & 35.89/5.10 \\
    & DBP+COP & 92.76 & \textbf{83.19} & \textbf{30.49}/\textbf{6.01} \\
    \midrule
    \multirow{8}{*}{C100}
    & R164 & \textbf{73.32} & 0.00 & 183.20/1.00 \\
    & Slim\cite{liu2017learning} & 73.25 & 42.88 & 130.59/1.40 \\
    & FPGM\cite{he2019filter} & 69.81 & 72.21 & 109.60/1.68 \\
    & COP\cite{wang2019cop} & 69.97 & 60.00 & 107.61/1.70 \\
    & DBP & 70.11 & 83.34 & 30.33/6.03 \\
    & DBP+Slim & 66.91 & \textbf{90.99} & 22.76/8.04 \\
    & DBP+FPGM & 67.03 & 90.96 & 18.60/9.85 \\
    & DBP+COP & 67.59 & 86.59 & \textbf{17.13}/\textbf{10.68} \\
    \bottomrule
\end{tabular}}
\end{center}
\caption{The results compared with filter-level pruning. ``Alg'' and ``Acc'' mean the algorithms we use and the accuracy of their pruned model, respectively. R164 means ResNet164, which is the un-pruned baseline model. ``Time'' and ``AR'' means the inference time in million second and acceleration ratio of pruned models. As we can see, compared with filter-level pruning, the inference of DBP-pruned model is at least 2.1 times as fast as those of other algorithms, and the inference of the combination of DBP and filter-level pruning is always the fastest.}
\label{tab:FP2}
\end{table}

\subsubsection{Results compared with filter-level pruning}

Slim~\cite{liu2017learning}, FPGM~\cite{he2019filter}, and COP~\cite{wang2019cop} are three state-of-the-art filter-level pruning algorithms, which respectively utilize different characteristics of filters to observe redundant filters. The experimental results are shown in Table~\ref{tab:FP2}. As we can see, though the FLOPs-reduction-ratio of DBP is smaller than filter-level pruning methods, its inference time is always much lower than filter-level pruning because of the pruning of layers. It is also worth highlighting that the combination of DBP and filter-level pruning algorithm achieves the highest acceleration ratio. Note that the inference of the COP-pruned model is faster than that of the FPGM-pruned model and the Slim-pruned model, even though COP reduces fewer FLOPs. The reason is that both FPGM~\cite{he2019filter} and Slim~\cite{liu2017learning} induce extra operations into the pruned model, which cost some inference time.

\subsection{Ablation study}

We ascribe the success of DBP to three aspects: 1) Pruning redundant blocks removes redundant information and preserves the important one. 2) Iterative pruning helps in advancing the accuracy. 3) MSE-based mimic loss provides useful information to the pruned model. In this section, we will explore the contributions of these aspects, and the experimental results are all in Table~\ref{tab:AB1}.

\begin{table*}[t]
\begin{center}
\begin{tabular}{c|ccccccc}
    \toprule
    Dataset & T-S & Un-pruned(\%) & DBP(\%) & Random(\%) & DBP-A(\%) & DBP-B(\%) & DBP-C(\%) \\
    \midrule
    \multirow{3}{*}{C10}
    & R164-R20 & 94.17 & 92.91 & 91.15$\pm$1.10(-1.76) & 92.74(-0.17) & 92.66(-0.25) & 91.82(-1.09) \\
    & R26-R8 & 92.55 & 89.69 & 87.45$\pm$1.02(-2.24) & 89.29(-0.40) & 89.75(+0.06) & 87.86(-1.83) \\
    & D100-D40 & 95.35 & 93.88 & 93.2$\pm$0.88(-0.68) & 91.53(-2.35) & 92.02(-1.86) & 91.15(-2.73) \\
    \midrule
    \multirow{3}{*}{C100}
    & R164-R20 & 73.32 & 69.67 & 66.76$\pm$2.63(-2.91) & 69.55(-0.12) & 69.31(-0.36) & 67.69(-1.98) \\
    & R32-R14 & 69.36 & 67.12 & 62.97$\pm$3.47(-4.15) & 66.93(-0.19) & 67.19(+0.07) & 65.00(-2.12) \\
    & D100-D40 & 76.52 & 72.65 & 71.80$\pm$0.77(-0.85) & 69.87(-2.78) & 71.32(-1.33) & 69.03(-3.62) \\
    \bottomrule
\end{tabular}
\end{center}
\caption{Ablation study. ``Random'' means pruning blocks randomly. ``DBP-A'' means pruning all redundant blocks once and fine-tuning without mimic loss. ``DBP-B'' means pruning all redundant blocks once and fine-tuning with mimic loss. ``DBP-C'' means pruning redundant blocks randomly and fine-tuning without mimic loss. The numbers in the brackets mean the accuracy changes compared with DBP.}
\label{tab:AB1}
\end{table*}

\subsubsection{Pruning blocks randomly}

To check the effectiveness of the discrimination based criterion, we compare it with a random pruning strategy. Specifically, we experiment under the same conditions with DBP except that the pruned blocks are chosen randomly. The experiment is repeated ten times, and the results are shown in the ``Random'' column of Table~\ref{tab:AB1}.

The results show that the average accuracy of a random pruning is always less than the accuracy of DBP, and the difference on CIFAR100 is bigger than on CIFAR10 because CIFAR100 is more complicated than CIFAR10. The results show that discrimination based pruning does remove redundant blocks and preserve important information. 

\subsubsection{Pruning all once \& fine-tuning without mimic-loss} \label{exp:abl:pao}

As described in Section~\ref{method:fine-tune}, two techniques are used when recovering the performance of our pruned model: iteratively pruning and MSE-based mimic loss. To check the enhancement brought by them, we use original DBP as our baseline and do three control experiments:

\begin{enumerate}[(a)]
\item Pruning all redundant blocks once \& fine-tuning without mimic loss. (DBP-A)
\item Pruning all redundant blocks once \& fine-tuning with mimic loss. (DBP-B)
\item Iteratively pruning the redundant blocks \& fine-tuning without mimic loss. (DBP-C)
\end{enumerate}


The results are shown in the ``DBP'', ``DBP-A'', ``DBP-B'', and ``DBP-C'' column of Table~\ref{tab:AB1}. Generally speaking, using both techniques achieves the highest accuracy, which means DBP is the best choice for both ResNet and DenseNet. However, as we can see, the impacts of these two techniques on ResNet and DenseNet are different:

\paragraph{Impacts on ResNet} The performance of DBP, DBP-A, and DBP-B is not much different. However, DBP-C yields awful results, which means the mimic loss is very important to iterative pruning. The results are kind of counter-intuitive. However, we think they are still reasonable. Specifically, if an intermediate-generated model does not converge well, it will influence the selection of blocks to be pruned, and thus affect the final accuracy of the pruned model, and mimic loss guarantees the fast convergence of intermediate-generated models.


\paragraph{Impacts on DenseNet} Only DBP yields satisfying results, which means both iterative pruning and mimic loss are essential when pruning DenseNet. The output features of a block in DenseNet are used by many other blocks, so pruning one block will affect many other blocks. In other words, DenseNet is more sensitive to pruning than ResNet, and pruning too many blocks once will lead to invocatable loss of performance.

\subsection{Case study}

\begin{figure}[t]
\begin{center}
\includegraphics[width=1.0\linewidth]{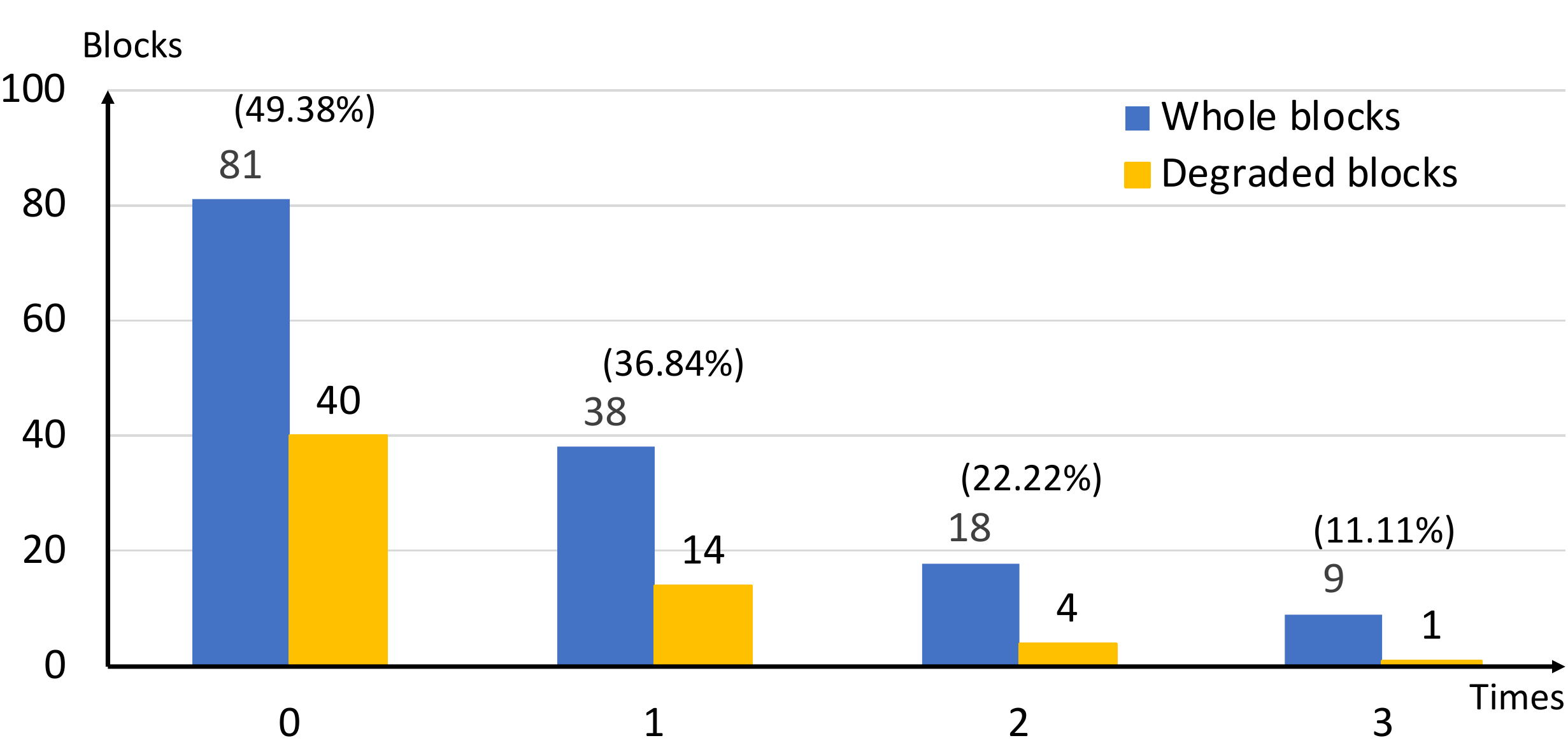}
\end{center}
\caption{The comparison between the number of whole blocks and that of degraded blocks. We count the number of whole blocks and degraded blocks of all models, \ie the un-pruned model and three pruned models. The percentages are the ratio of redundant blocks to the whole blocks. As we can see, the number of degraded blocks decreases faster than the whole number of blocks as the pruning progresses.}
\label{fig:cs}
\end{figure}

In this section, we explore what happens during the pruning process. It is worth recalling that we defined a degraded block as the block at which the discrimination of features degrades. We prune ResNet with 81 blocks to 9 blocks through three times of pruning. As shown in Figure~\ref{fig:cs}, the ratio of redundant blocks in the whole blocks decreases as pruning progresses.

\begin{figure}[t]
\begin{center}
\includegraphics[width=1.0\linewidth]{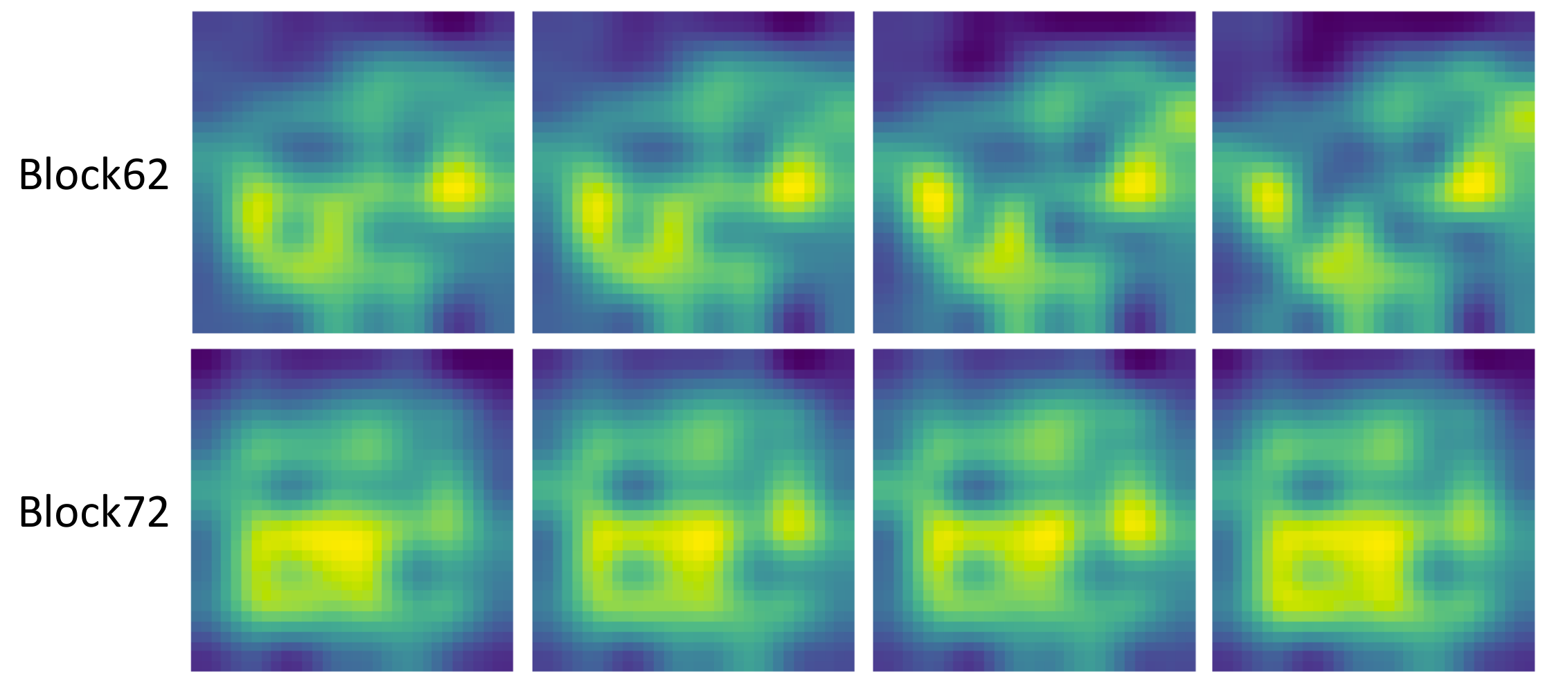}
\end{center}
\caption{The output features of preserved blocks. Block62 and Block72 are two of preserved blocks after pruning. We take a random image in CIFAR10 as input and visualize the output of Block62 and Block72 for four models, \ie, the un-pruned model and three pruned models. The output features of these two blocks look similar before and after pruning.}
\label{fig:cs2}
\end{figure}

We also track some preserved blocks after pruning, the Figure~\ref{fig:cs2} shows that the output features of preserved blocks look similar before and after pruning, which means that DBP preserves the important information, and the pruned model utilizes fewer blocks to express similar information to the original model.

\section{Conclusion}

In this paper, we point out the limited performance of filter-level pruning in accelerating the inference of CNNs and analyze the reasons. To solve the problem, a discrimination based block-level pruning (DBP) is proposed. DBP outperforms the state-of-the-art and achieves a considerable acceleration ratio with acceptable accuracy loss.

{\small
\bibliographystyle{ieee_fullname}
\bibliography{egbib1}
}

\end{document}